\pdfoutput=1
\documentclass[11pt]{article}

\usepackage{ACL2023}

\usepackage{times}
\usepackage{latexsym}

\usepackage[T1]{fontenc}
\usepackage[utf8]{inputenc}

\usepackage{microtype}

\usepackage{inconsolata}

\usepackage{graphicx} 
\usepackage{caption,subcaption}
\usepackage{amsmath}
\usepackage{xspace}
\usepackage{nccmath}
\usepackage{amsfonts} 
\usepackage{booktabs} 
\usepackage[ruled,noend]{algorithm2e}
\DontPrintSemicolon

\newcommand{\renet}{\textsc{RE-Net}\xspace}
\newcommand{\hdb}{HDBSCAN\xspace}

\newcommand{\datam}{ICEWS-M\xspace}
\newcommand{\datatm}{ICEWS-2M\xspace}
\newcommand{\gdelt}{GDELT\xspace}

\newcommand{\prob}{p}
\newcommand{\bs}{\mathrm{s}}
\newcommand{\br}{\mathrm{r}}
\newcommand{\bo}{\mathrm{o}}
\newcommand{\bw}{\boldsymbol{w}}
\newcommand{\bh}{\boldsymbol{h}}
\newcommand{\es}{\boldsymbol{e}_{\mathrm{s}}}
\newcommand{\er}{\boldsymbol{e}_{\mathrm{r}}}
\newcommand{\eo}{\boldsymbol{e}_{\mathrm{o}}}
\newcommand{\mbR}{\mathbb{R}}

\newcommand{\calT}{\mathcal{T}}
\newcommand{\calM}{\mathcal{M}}
\newcommand{\calB}{\mathcal{B}}
\newcommand{\calC}{\mathcal{C}}
\newcommand{\calX}{\mathcal{X}}
\newcommand{\tr}{D^{train}}
\newcommand{\te}{D^{test}}
\newcommand{\val}{D^{val}}

\title{History Repeats: Overcoming Catastrophic Forgetting For Event-Centric Temporal Knowledge Graph Completion}

\author{Mehrnoosh Mirtaheri \quad Mohammad Rostami \quad Aram Galstyan \\
        Information Sciences Institute \quad University of Southern California\\
        \texttt{mirtaheri@usc.edu \quad \{rostami, galstyan\}@isi.edu} \\}
\begin{document}
\maketitle
\begin{abstract}
Temporal knowledge graph (TKG) completion models typically rely on having access to the entire graph during training. However, in real-world scenarios, TKG data is often received incrementally as events unfold, leading to a dynamic non-stationary data distribution over time. While one could incorporate fine-tuning to existing methods to allow them to adapt to evolving TKG data, this can lead to forgetting previously learned patterns. Alternatively, retraining the model with the entire updated TKG can mitigate forgetting but is computationally burdensome. To address these challenges, we propose a general continual training framework that is applicable to any TKG completion method, and leverages two key ideas: (i) a temporal regularization that encourages repurposing of less important model parameters for learning new knowledge, and (ii) a clustering-based experience replay that reinforces the past knowledge by selectively preserving only a small portion of the past data. Our experimental results on widely used event-centric TKG datasets demonstrate the effectiveness of our proposed continual training framework in adapting to new events while reducing catastrophic forgetting. Further, we perform ablation studies to show the effectiveness of each component of our proposed framework. Finally, we investigate the relation between the memory dedicated to experience replay and the benefit gained from our clustering-based sampling strategy. 

\end{abstract}

\section{Introduction}
Knowledge graphs (KGs) provide a powerful tool for studying the underlying structure of multi-relational data in the real world \cite{liang2022reasoning}. They present factual information in the form of triples, each consisting of a subject entity, a relation, and an object entity. Despite the development of advanced extraction techniques, knowledge graphs often suffer from incompleteness, which can lead to errors in downstream applications. As a result, the task of predicting missing facts in knowledge graphs, also known as knowledge graph completion, has become crucial. \cite{wang2022simkgc,huang2022multilingual,shen2022comprehensive}

KGs are commonly extracted from real-world data streams, such as newspaper texts that change and update over time, making them inherently dynamic. The stream of data that emerges every day may contain new entities, relations, or facts. As a result, facts in a knowledge graph are usually accompanied by time information. A fact in a semantic knowledge graph, such as Yago~\cite{kasneci2009yago}, may be associated with a time interval, indicating when it appeared and remained in the KG. For example, consider \textit{(Obama, President, United States, 2009-2017)} in a semantic KG. A link between \textit{Obama} and \textit{United states} appears in the graph after 2009, and it exists until 2017. On the other hand, a fact in a Temporal event-centric knowledge graph (TKGs), such as ICEWS~\cite{boschee2015integrated}, is associated with a single timestamp, indicating the exact time of the interaction between the subject and object entities. For example, in an event-centric TKG, \textit{(Obama, meet, Merkel)} creates a link between \textit{Obama} and \textit{Merkel} several times within 2009 to 2017 since the temporal links only show the time when an event has occurred. Therefore, event-centric TKGs exhibit a high degree of dynamism and non-stationarity in contrast to semantic KGs.

\begin{figure}[t!]
    \centering
    \includegraphics[width=.48\textwidth]{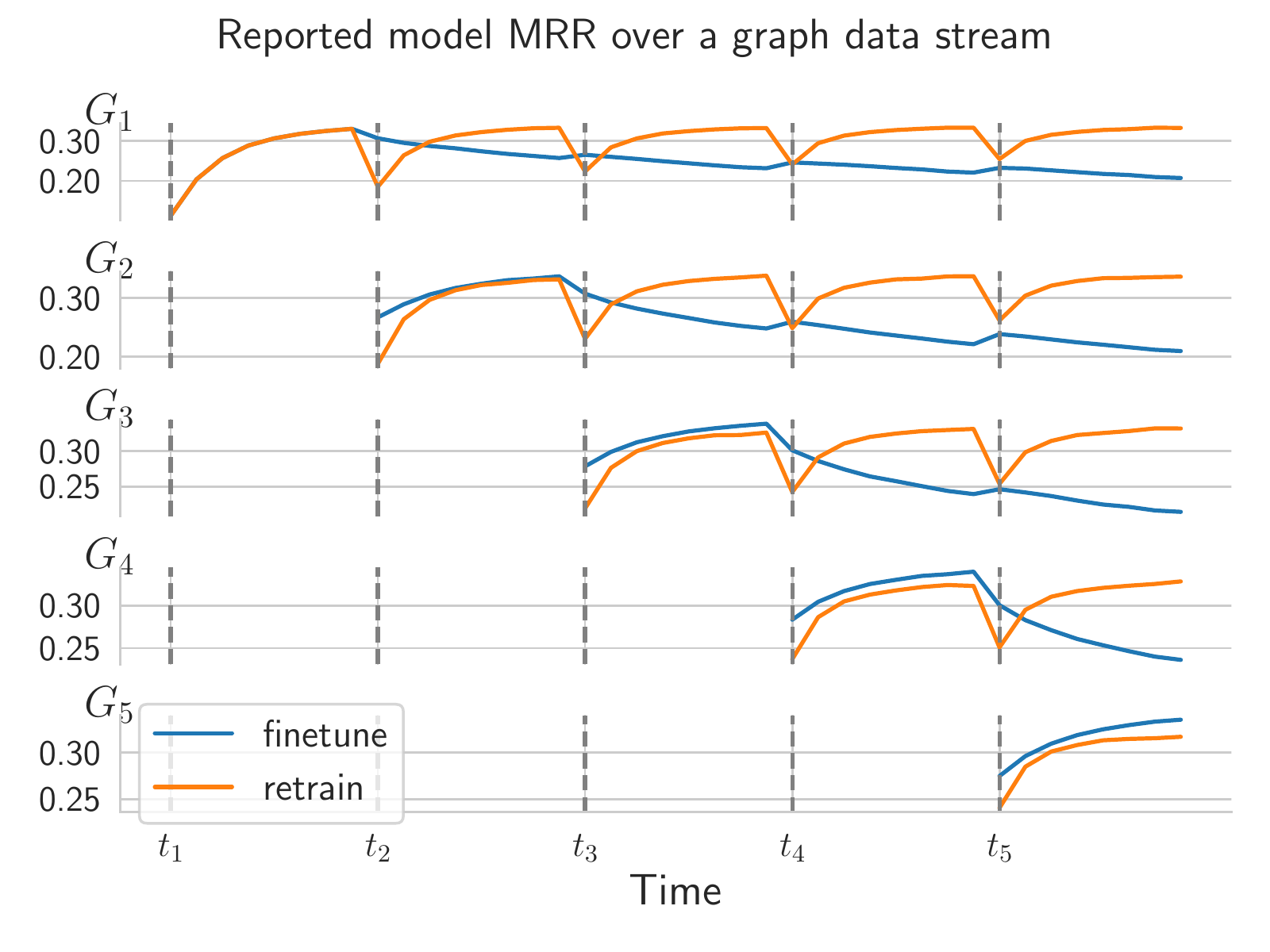}
    \caption{Catastrophic forgetting effect of fine-tuning. A TKG completion model is fine-tuned with the graph data at time $t_i$ and achieves the highest MRR score for $G_i$. The MRR scores decrease for $G_{1}, ..., G_{i-1}$.}
    \label{fig:forgetting}
    \vspace{-.5cm}
\end{figure}

To effectively capture the temporal dependencies within entities and relations in TKGs, as well as new patterns that may emerge with new data streams, it is necessary to develop models specifically designed for TKG completion. A significant amount of research has been dedicated to developing evolving models~\cite{Messner2022, mirtaherione, jin2020recurrent,garg2020temporal} for TKG completion. These models typically assume evolving vector representations for entities or relations. These representations change depending on the timestep, and they can capture temporal dependencies between entities. However, these models often assume that the entire dataset is available during training. They do not provide a systematic method for updating model parameters when new data is added. One potential solution is to retrain the model with new data. However, this approach can be resource-intensive and impractical for large-scale knowledge graphs. An alternative approach is to fine-tune the model with new data, which is more time and memory efficient. However, this approach has been shown to be susceptible to overfitting to the new data, resulting in the model forgetting previously learned knowledge, a phenomenon known as catastrophic forgetting (Fig. \ref{fig:forgetting}).  A limited number of studies \cite{Song2018,Daruna2021,Wu2021} have addressed this problem for semantic knowledge graphs using continual learning approaches, with TIE \cite{Wu2021} being the most closely related work to current research. Nevertheless, the development of efficient and effective methods for updating models with new data remains a significant challenge in event-centric Temporal Knowledge Graphs.

We propose a framework for incrementally training a TKG completion model that consolidates the previously learned knowledge while capturing new patterns in the data. Our incremental learning framework employs regularization and experience replay to alleviate catastrophic forgetting. We propose a temporal regularization method based on elastic weight consolidation \cite{kirkpatrick2017overcoming}. By estimating an importance weight for every model parameter at each timestep, the regularization term in the objective function 'freezes' the more important parameters from past timesteps, encouraging the use of less important parameters for learning the current task. Additionally, an exponentially decaying hyperparameter in the objective function further emphasizes the importance of the most recent tasks over older ones. Our selective experience replay method uses clustering over the representation of the data points to first capture the underlying structure of the data. The points closest to the clusters' centroid are selected for experience replay. We show that the temporal regularization combined with clustering-based experience replay outperforms all the baselines in alleviating catastrophic forgetting. Our main contributions include:
\begin{enumerate}
    \item A novel framework for incremental training and evaluation of event-centric TKGs, which addresses the challenges of efficiently updating models with new data.
    \item A clustering-based experience replay method, which we show to be more effective than uniform sample selection. We also demonstrate that careful data selection for experience replay is crucial when memory is limited. 
    \item An augmentation of the training loss with a consolidation loss, specifically designed for TKG completion, which helps mitigate forgetting effects. We show that assigning a decayed importance to the older tasks reduces forgetting effects. 
    \item A thorough evaluation of the proposed methods through extensive quantitative experiments to demonstrate the effectiveness of our full training strategies compared to baselines.
\end{enumerate}
\section{Related Work}
Our work is related to TKG completion, continual learning methods, and recent developments of continual learning for knowledge graphs. 

\subsection{Temporal Knowledge Graph Reasoning} TKG completion methods can be broadly categorized into two main categories based on their approach for encoding time information: translation-based methods and evolving methods.

Translation-based methods, such as those proposed by \cite{leblay2018deriving, garcia2018learning, dasgupta2018hyte, wang2019hybrid, jain2020temporal}, and \cite{Sadeghian2021}, utilize a lower-dimensional space, such as a vector~\cite{leblay2018deriving,jain2020temporal}, or a hyperplane~\cite{dasgupta2018hyte,wang2019hybrid}, for event timestamps and define a function to map an initial embedding to a time-aware embedding.

On the other hand, evolving models assume a dynamic representation for entities or relations that is updated over time. These dynamics can be captured by shallow encoders~\cite{xu2019temporal,mirtaheri2019tensor,han2020dyernie} or sequential neural networks~\cite{trivedi2017know,jin2020recurrent,wu2020temp,zhu2020learning,han2020xerte,han2020graph,Li2021kg}. For example,\cite{xu2019temporal} model entities and relations as time series, decomposing them into three components using adaptive time series decomposition. DyERNIE~\cite{han2020dyernie} propose a non-Euclidean embedding approach in the hyperbolic space. \cite{trivedi2017know} represent events as point processes, while ~\cite{jin2020recurrent} utilizes a recurrent architecture to aggregate the entity neighborhood from past timestamps.

\subsection{Continual Learning}
Continual learning (CL) or lifelong learning is a learning setting where a set of tasks are learned in a sequence. The major challenge in CL is overcoming catastrophic forgetting, where the model's performance on past learned tasks is degraded as it is updated to learn new tasks in the sequence. Experience replay~\cite{li2018learning} is a major approach to mitigate forgetting, where representative samples of past tasks are replayed when updating a model to retain past learned knowledge. To maintain a memory buffer storage with a fixed size, representative samples must be selected and discarded. \cite{schaul2015prioritized} propose selecting samples that led to the maximum effect on the loss function when learning past tasks.

To relax the need for a memory buffer, generative models can be used to learn generating pseudo-samples. \cite{shin2017continual} use adversarial learning for this purpose. An alternative approach is to use data generation using autoencoders\cite{rostami2020generative,rostami2023drifting}. Weight consolidation is another important approach to mitigate catastrophic forgetting~\cite{zenke2017temporal,kirkpatrick2017overcoming}. The   idea is to identify important weights that play an important role in encoding the learned knowledge about past tasks and consolidate them when the model is updated to learn new tasks. As a result, new tasks are learned using primarily the free learnable weights. In our framework, we combine both approaches to achieve optimal performance.

\subsection{Continual Learning for Graphs}
CL in the context of graph structures remains an under-explored area, with a limited number of recent studies addressing the challenge of dynamic heterogeneous networks~\cite{Tang2020, wang2020streaming,zhou2021overcoming} and semantic knowledge graphs~\cite{Song2018,Daruna2021,Wu2021}. In particular, \cite{Song2018, Daruna2021} propose methods that integrate class incremental learning models with static translation-based approaches, such as TransE~\cite{bordes2013translating}, for addressing the problem of continual KG embeddings. Additionally, TIE \cite{Wu2021} develops a framework that predominantly focuses on semantic KGs, and generates yearly graph snapshots by converting a fact with a time interval into multiple timestamped facts. This process can cause a loss of more detailed temporal information, such as the month and date, and results in a substantial overlap of over 95\% between consecutive snapshots. TIE's frequency-based experience replay mechanism operates by sampling a fixed set of data points from a fixed-length window of past graph snapshots; for instance, at a given time $t$, it has access to the snapshots from $t-1$ to $t-5$. This contrasts with the standard continual learning practice, which involves sampling data points from the current dataset and storing them in a continuously updated, fixed-size memory buffer. When compared to Elastic Weight Consolidation (EWC), the L2 regularizer used by TIE proves to be more rigid when learning new tasks over time. Furthermore, their method's evaluation is confined to shallow KG completion models like Diachronic Embeddings \cite{goel2020diachronic} and HyTE \cite{dasgupta2018hyte}.

\section{Problem Definition}
This section presents the formal definition of continual temporal knowledge graph completion. 
\subsection{Temporal Knowledge Graph Reasoning}
A TKG is a collection of events represented as a set of quadruples $G = \{(s, r, o, \tau) | s, o \in \mathcal{E}, r \in \mathcal{R}\}$, where $\mathcal{E}$ and $\mathcal{R}$ are the set of entities and relations, and $\tau$ is the timestamp of the event occurrence. These events represent one-time interactions between entities at a specific time. The task of temporal knowledge graph completion is to predict whether there will be an interaction between two entities at a given time. This can be done by either predicting the object entity, given the subject and relation at a certain time, or by predicting the relation between entities, given the subject and object at a certain time. In this case, we will focus on the first method which can be formally defined as a ranking problem. The model will assign higher likelihood to valid entities and rank them higher than the rest of the candidate entities.

\subsection{Continual Learning Framework For Tempporal Knolwedge Graphs}
\label{secs:cl_setup}
A Temporal knowledge graph $G$ can be represented as a stream of graph snapshots $G_1, G_2, \dots, G_T$ arriving over time,  where $G_t = \{(s, r, o, \tau) | s, o \in \mathcal{E}, r \in \mathcal{R}, \tau \in \left[ \tau_t, \tau_{t+1} \right) \}$ is a set of events occurred within time interval $\left[\tau_t, \tau_{t+1}\right)$.

The continual training of a TKG completion method involves updating the parameters of the model $\mathcal{M}$ as new graph snapshots, consisting of a set of events, become available over time. This process aims to consolidate previously acquired information while incorporating new patterns. Formally, we define a set of tasks $\langle\mathcal{T}_1, \dots, \mathcal{T}_T \rangle$, where each task $\mathcal{T}_t = \left(D_t^{train}, D_t^{test}, D_t^{val}\right)$ is comprised of disjoint subsets of the $G_t$ events, created through random splitting. A continually trained  model $\mathcal{M}$ can then be shown as a stream of models $\calM = \langle\calM_1, \dots, \calM_T\rangle$, with corresponding parameter sets $\theta = \langle \theta_1, \theta_2, ..., \theta_T\rangle$, trained incrementally as a stream of tasks arrive $\calT = \langle\calT_1, \calT_2, ..., \calT_T\rangle$.

\subsection{Base Model}
In this paper, we utilize \renet~\cite{jin2020recurrent}, a state-of-the-art TKG completion method, as the base model. \renet is a recurrent architecture for predicting future interactions, which models the probability of an event occurrence based on temporal sequences of past knowledge graphs. The model incorporates a recurrent event encoder to process past events and a neighborhood aggregator to model connections at the same time stamp. Although \renet was initially developed for predicting future events (extrapolation), it can also be used to predict missing links in the current state of the graph (interpolation), which is the focus of this study. The model parameterizes the probability of an event $\prob(\bo_\tau| \bs,\br)$ as follows:

\begin{equation}
\small \prob( \bo_{\tau}|\bs,\br) \propto \exp \left([\es:\er:\bh_{\tau-1}(\bs, \br)]^\top \cdot \bw_{\bo_{\tau}}\right),
\end{equation}

where $\es, \er \in \mbR^d$ are learnable embedding vectors for the subject entity $\bs$ and relation $\br$. $\bh_{\tau-1}(\bs,\br)\in \mbR^d$ represents the local dynamics within a time window $(\tau-\ell, \tau-1)$ for $(\bs,\br)$. By combining both the static and dynamic representations, \renet effectively captures the semantics of $(\bs,\br)$ up to time stamp $(\tau-1)$. The model then calculates the probability of different object entities $\bo_{\tau}$ by passing the encoding through a multi-layer perceptron (MLP) decoder, which is defined as a linear softmax classifier parameterized by ${\bw_{\bo_{\tau}}}$.

\section{Methodology}
Our proposed framework is a training approach that can be applied to any TKG completion model. It enables the incremental updating of model parameters with new data while addressing the issues of catastrophic forgetting associated with fine-tuning. To achieve this, we utilize experience replay and regularization techniques - methodologies commonly employed in image processing and reinforcement learning to mitigate forgetting. Additionally, we introduce a novel experience replay approach that employs clustering to identify and select data points that best capture the underlying structure of the data. Furthermore, we adopt the regularization method of EWC, as proposed in [Kirkpatrick et al., 2017], which incorporates a decay parameter that assigns higher priority to more recent tasks. Our results demonstrate that the incorporation of a decay parameter into the EWC loss and prioritizing more recent tasks leads to improved performance.
\subsection{Experience Replay}
In the field of neuroscience, the hippocampal replay, or the re-activation of specific trajectories, is a crucial mechanism for various neurological functions, including memory consolidation. Motivated by this concept, the use of experience replay in Continual Learning (CL) for deep neural networks aims to consolidate previously learned knowledge when a new task is encountered by replaying previous experiences, or training the model on a limited subset of previous data points. However, a challenge with experience replay, also known as memory-based methods, is the requirement for a large memory size to fully consolidate previous tasks~\cite{rostami2023overcoming}. Thus, careful selection of data points that effectively represent the distribution of previous data becomes necessary.

In this work, we propose the use of experience replay for continual TKG completion. Specifically, we maintain a memory buffer $\calB$ which, at time $t$, contains a subset of events sampled from $\tr_1, \tr_2, \dots, \tr_{t-1}$. When Task $\calT_t$ is presented to the model, it is trained on the data points in $\tr_t \cup \calB$. After training, a random subset of events in the memory buffer, $\frac{|\calB|}{t}$, are discarded and replaced with a new subset of events sampled from $\tr_{t}$. In this way, at time $t$, where $t$ tasks have been observed, equal portions of memory with size $\frac{|\calB|}{t}$ are dedicated to each task. A naive approach for selecting a subset of events from a task's training set at time $t$ would be to uniformly sample $\frac{|\calB|}{t}$ events from $\tr_t$. However, we propose a clustering-based sampling method that offers a more careful selection algorithm, which is detailed in the following section.
\subsubsection{Clustering-based Sampling}
\label{sec:cer}
When dealing with complex data, it is likely that various subspaces exist within the data that must be represented in the memory buffer. To address this issue, clustering methods are employed to diversify the memory buffer by grouping data points into distinct clusters. The centroids of these clusters can be utilized as instances themselves or as representatives of parts of the memory buffer.\cite{shi2018sample,hayes2019memory,korycki2021class}. In this study, clustering is applied to the representation of events in the training set in order to uncover the underlying structure of the data and select data points that effectively cover the data distribution. The Hierarchical Density-Based Spatial Clustering of Applications with Noise (\hdb) algorithm \cite{mcinnes2017hdbscan} is utilized for this purpose. \hdb is a hierarchical, non-parametric, density-based clustering method that groups points that are closely packed together while identifying points in low-density regions as outliers. 

The use of \hdb over other clustering methods is advantageous due to its minimal requirements for hyperparameters. Many clustering algorithms necessitate and are sensitive to the number of clusters as a hyperparameter. However, \hdb can determine the appropriate number of clusters by identifying and merging dense space regions. Additionally, many clustering algorithms are limited to finding only spherical clusters. \hdb, on the other hand, is capable of uncovering more complex underlying structures in the data. As a result of its ability to identify clusters with off-shaped structures, \hdb generates a set of exemplar points for each cluster rather than a single point as the cluster centroid.

We represent each event $(s, r, o, \tau) \in \tr_t$ as a vector $[\es: \eo] \in \mbR^{2d}$, where $\es$ and $\eo$ represent the $d$-dimensional embeddings of $s$ and $o$ at time $t$, respectively. The notation [:] denotes concatenation, creating a $|\tr_t| \times 2d$ matrix that represents the training data at time $t$. In our initial experiments, we found that data representations such as $[\es:\er]$, where $\er$ is the relation embeddings,  did not significantly affect the results. Moreover, representing the data as $[\es:\er:\eo]$ led to a bias towards relation representation, causing data points with identical relation types to cluster together.

We obtained clusters $\calC^1, \calC^2, \dots, \calC^m$ by running \hdb. Our algorithm then selects $\frac{|\calB|}{t}$ events from these clusters by prioritizing the data points closest to the exemplars and giving precedence to larger clusters. If $\frac{|\calB|}{t} < m$, data points are chosen only from the first $\frac{|\calB|}{t}$ clusters. Conversely, if $\frac{|\calB|}{t} > m$, the number of points selected from each cluster will depend on the cluster size, with a minimum of one data point chosen from each cluster. The specifics of this procedure are detailed further in Algorithm \ref{alg:er_select}.

\newlength{\origtextfloatsep}
\setlength{\origtextfloatsep}{\textfloatsep}
\setlength{\textfloatsep}{3pt}
\SetKwInput{KwData}{input}
\SetKwInput{KwResult}{output}
\begin{algorithm}[t]
%\begin{algorithmic}
\SetAlgoLined
\definecolor{CommentColor}{HTML}{006619}
\KwData{
    $\calC_t = \calC^1_t, \calC^2_t, \dots, \calC^m_t$ \textcolor{CommentColor}{(clusters generated with hdbscan from $\tr_t$ sorted in decreasing order of their size};
    $\tr_t$ \textcolor{CommentColor}{(training set at time $t$)};
    $s$ \textcolor{CommentColor}{(sample size)}; 
    $\textsc{\texttt{FindExemplars}}(\calC^i, k)$ (\textcolor{CommentColor}{Takes a cluster and returns $k$ points closests to the cluster exemplars.)}
    }
%\KwResult{Write here the result }
  \SetKwProg{Fn}{def}{:}{}
  \nl
  \Fn{SelectPoints($\calC_t$, $\tr_t$, $s$)}
  {
  \nl
  $Q \leftarrow \emptyset$ \;
  \nl
  \For{$i \leftarrow 1$ \textbf{to} $m$ }{
  \nl
  $r \leftarrow \lceil\frac{|\calC^i|}{\sum_j |\calC^j|}\times s \rceil $\;
  \nl
  $\calX \leftarrow \textsc{\texttt{FindExemplars}}(\calC^i, r)$\;
  \nl
  $Q \leftarrow Q \cup (\mathcal{X}, r)$
  }
 \nl
  $S \leftarrow \emptyset$\;
  \nl 
\While {$Q \neq \emptyset \And |S| < s$} {
\nl
$\calX, r \leftarrow Q.\texttt{pop}()$\;
\nl
$S \leftarrow S \cup [\calX[0]]$\;
\nl
$Q \leftarrow Q \cup (\calX[1:], r-1)$
}
\nl
  \textbf{return} $S$
}
 \caption{Cluster Experience Replay}
\label{alg:er_select}
\end{algorithm}

\subsection{Regularization}
Regularization-based approaches for CL incorporate a regularization term in the objective function to discourage changes in the weights that are crucial for previous tasks, while encouraging the utilization of other weights. One such approach, Elastic Weight Consolidation (EWC) \cite{kirkpatrick2017overcoming}, estimates the importance of weights using the Fisher Information Matrix. Given a model with parameter set $\theta$ previously trained on task $A$, and a new task B, EWC optimizes the following loss function:
\begin{equation}
    \mathcal{L}(\theta) = \mathcal{L}_B(\theta) + \sum_i \frac{\lambda}{2} F_i (\theta_i - \theta^*_{A, i})^2
\end{equation}

Where $\mathcal{L}_B$ is the loss over task $B$ only and $\lambda$ determines the importance of the previous task compared to task $B$. We extend this loss function for continual TKG completion. Given a stream of tasks $\langle \calT_1, \calT_2 \dots, \calT_t \rangle$ and incrementally obtained parameter sets $\langle\theta_1, \theta_2 \dots, \theta_t\rangle$, we define the temporal EWC loss functions as follows:

\begin{equation}
    \label{eq:temp_ewc}
    \mathcal{L}(\theta_t) = \mathcal{L}_{\calT_t}(\theta_t) + \sum_{\tau=1}^{t-1} \sum_i \frac{\lambda}{2} F_\tau (\theta_i - \theta^*_{\tau, i})^2
\end{equation}

Where $\mathcal{L}_{\calT_t}$ is the model loss calculated only using $\calM_t$ and $\tr_t$, $F_\tau$ is the Fisher Information Matrix estimated for $\calM_\tau$ and $\calT_\tau$ and $\theta_{\tau, i}$ is $i$-th parameter of $\calM_\tau$. The $\lambda$ parameter in Equation \ref{eq:temp_ewc} assigns equal importance to all the tasks from previous time steps, however, in practice, and depending on the application, different tasks might have different effect on the current task making plausibility of   adaptive $\lambda_\tau$:

\begin{equation}
    \label{eq:temp_ewc_lambda}
    \mathcal{L}(\theta_t) = \mathcal{L}_{\calT_t}(\theta_t) + \sum_{\tau=1}^{t-1} \sum_i \frac{\lambda_\tau}{2} F_\tau (\theta_i - \theta^*_{\tau, i})^2,
\end{equation}
where $\lambda_\tau = \lambda \alpha ^ {t-\tau}$, $\lambda$ is the overall EWC loss importance, and $\alpha < 1$ is the decay parameter.

\subsection{Training and Loss Function}
The final loss function of our framework, when trained with experience replay and EWC can be summarized as follows: 

\begin{equation}
    \begin{split}
        &\mathcal{L}(\theta_t) = \mathcal{L}_{expr}(\theta_t) + \lambda\mathcal{L}_{ewc}(\theta_t),\\
        &\mathcal{L}_{expr}(\theta_t) = \mathcal{L}_{\calT_t \cup \calB}(\theta_t),\\
        &\mathcal{L}_{ewc} = \sum_{\tau=1}^{t-1} \sum_i \frac{ \alpha ^ {t-\tau}}{2} F_\tau (\theta_i - \theta^*_{\tau, i})^2
    \end{split}
\end{equation}

The replay loss $\mathcal{L}_{expr}$ is the model loss trained over both the current task's training set $\tr_t$ and the data points in the memory buffer $\calB$. For training in batches, the number of data points selected from $\tr_t$ and $\calB$ is in proportion to their size.

\section{Experiments}
\begin{figure*}[t!]
    \centering
    \includegraphics[width=\textwidth, trim={3cm 0 3cm 0}, clip]{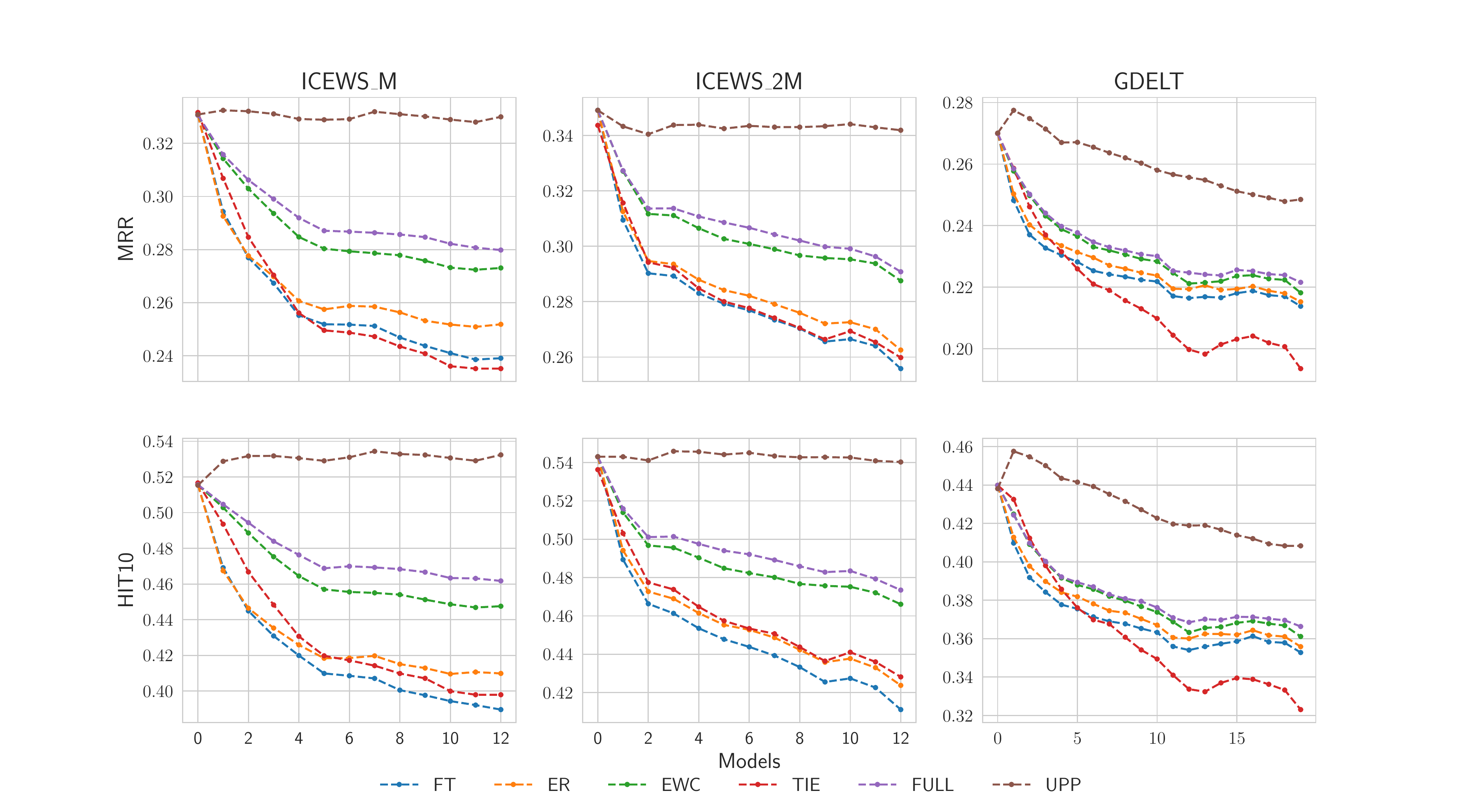}
    \caption{The overall performance comparison. Average Hit@10 and average MRR reported for \renet incrementally trained using three benchmarks: \datam, \datatm, and \gdelt.}
    \label{fig:fig:main}
    \vspace{-.5cm}
\end{figure*}
In this section, we explain the evaluation protocol to quantitatively measuring the model catastrophic forgetting. From know TKG datasets, we create two benchmarks for TKG continual learning. We evaluate our proposed training method using the benchmark, compare them with various baselines and show the effectiveness of our approach in alleviating catastrophic forgetting. Finally, we conduct ablation studies on different components of our training method to validate our model. 
\subsection{Datasets}
\begin{table}
\small
\setlength\tabcolsep{3pt}
    \centering
    \begin{tabular}{c|c c c  c c}
    \toprule
        Dataset & \#tasks & task period & split ratio & \shortstack{avg \#quads \\ train/test} & \\
        \midrule
        \datam & 13 & 1 month & 50/25/25 & 27k/13k  \\
        \datatm & 13 & 2 month & 50/25/25 & 50k/25k \\
        \gdelt & 21 & 3 days & 60/20/20 & 38k/13k \\
    \bottomrule
    \end{tabular}
    \caption{Dataset statistics}
    \label{tab:data}
    
\end{table}
We use two datasets: the Integrated Crisis Early Warning System (ICEWS) and the Global Database of Events, Language, and Tone (GDELT). Both datasets contain interactions between geopolitical actors, with daily event dates in the ICEWS dataset and 15-minute intervals in the GDELT dataset. To create benchmarks, we use a one-year period of the ICEWS dataset starting from 01-01-2015 and consider each month as a separate graph snapshot (\datam). We also use a two-year period from 01-01-2015 to 02-01-2017, dividing it into 13 graph snapshots with 2-month windows (\datatm). We split the events in each snapshot into train, validation, and test sets with a 50/25/25  percent ratio. For the GDELT, we use a 20-day period, dividing it into 3-day windows and split the data into train/test/validation sets with a 60/20/20 percent ratio. Table \ref{tab:data} includes statistics for each benchmark. We assume that all relations and entities are known at all times during training, and no new entities or relations are presented to the model.
\subsection{Evaluation Setup}
% As explained in Section \ref{secs:cl_setup}, given a model $\calM$ with parameter set $\theta$, we incrementally train the model using a stream of tasks $\calT_1, \calT_2, \dots, \calT_t$. Each task $\calT_tau$ comprises of a training , validation and test set denoted as $\tr_\tau$, $\val_\tau$, $\te_\tau$. 
We start by training $\calM$ over $\tr_1$ and use $\val_1$ for hyper-parameter tuning. The model $\calM_t$ with parameter set $\theta_t$ at time step $t$ is first initialized with parameters from the previous time step $\theta_{t-1}$. Then $\calM_t$  parameters are updated by training the model over $\tr_t$. The training step can be a simple fine-tuning, or it can be augmented with data points for experience replay or with the temporal EWC loss.

In order to assess the forgetting effect, at time $t$, we report the average  $\calM_t$ performance over the current and all the previous test sets $\te_1, \te_2, \dots, \te_t$. Precisely, we report the performance at time $t$ as $P_t = \frac{1}{t} \sum_{j=1}^t p_{t, j}$, where $p_{t, j}$ is the performance of $\calM_t$ measured by either MRR or Hit@10 over $\te_j$.

\subsection{Comparative Study}
To evaluate the performance of our incremental training framework, we conduct a comparative analysis with several baseline strategies. These include:

\begin{itemize}
\setlength\itemsep{0em}
\item \textbf{FT}: This strategy fine-tunes the model using the original loss function and the newly added data points.
\item \textbf{ER}: This method applies experience replay \cite{rolnick2019experience} with randomly chosen points. It then fine-tunes the model with both newly added events and events stored in the memory buffer.
\item \textbf{EWC} \cite{kirkpatrick2017overcoming}: In this strategy, the model is trained with a loss function augmented by an EWC (Elastic Weight Consolidation) loss, as defined in Equation \ref{eq:temp_ewc}.
\item \textbf{TIE} \cite{Wu2021}: Drawing from TIE's methodology, we incorporated L2 regularization into our objective function and utilized their implementation of frequency-based experience replay.
\item \textbf{Full}: Our comprehensive model is trained using a clustering-based experience replay mechanism, supplemented with a decayed EWC loss.
\end{itemize}
\begin{table*}[t]
\centering
\small
\setlength\tabcolsep{2pt}
    \begin{tabular}{l | c c c c | c c c c | c c c c}
        \toprule
        \multicolumn{1}{c}{} & \multicolumn{4}{c}{\datam} & \multicolumn{4}{c}{\datatm} & \multicolumn{4}{c}{\gdelt}\\
        \cmidrule(lr){2-5}
        \cmidrule(lr){6-9}
        \cmidrule(lr){10-13}
        \multicolumn{1}{c}{} & \multicolumn{2}{c}{Current} & \multicolumn{2}{c}{Average} & \multicolumn{2}{c}{Current} & \multicolumn{2}{c}{Average} & \multicolumn{2}{c}{Current} & \multicolumn{2}{c}{Average} \\
        \cmidrule(lr){2-3}
        \cmidrule(lr){4-5}
        \cmidrule(lr){6-7}
        \cmidrule(lr){8-9}
        \cmidrule(lr){10-11}
        \cmidrule(lr){11-13}
        Model & H@10 & MRR & H@10 & MRR & H@10 & MRR & H@10 & MRR & H@10 & MRR & H@10 & MRR\\
        \midrule
FT & $.503$& $.325$& $.390 \pm.04$& $.239 \pm.03$& $.517$& $.330$& $.411 \pm.04$& $.256 \pm.03$& $.421$& $.260$& $.351 \pm.02$& $.214 \pm.02$ \\ 
ER & $.491$& $.314$& $.410 \pm.03$& $.252 \pm.02$& $.521$& $.331$& $.424 \pm.04$& $.263 \pm.03$& $.429$& $.263$& $.356 \pm.02$& $.215 \pm.02$ \\ 
EWC & $.483$& $.299$& $.448 \pm.03$& $.273 \pm.02$& $.475$& $.294$& $.466 \pm.02$& $.288 \pm.02$& $.429$& $.262$& $.359 \pm.02$& $.217 \pm.02$ \\ 
TIE & $.548$& $.354$& $.398 \pm.05$& $.235 \pm.04$& $.567$& $.362$& $.428 \pm.05$& $.260 \pm.04$& $.492$& $.309$& $.322 \pm.05$& $.192 \pm.03$ \\ 
\midrule
OURS & $.565$& $.358$& \textbf{$\mathbf{.462 \pm.04}$}& $\mathbf{.280 \pm.03}$& $.555$& $.349$& $\mathbf{.473 \pm.03}$& $\mathbf{.291 \pm.02}$& $.416$& $.256$& $\mathbf{.365 \pm.02}$& $\mathbf{.222 \pm.01}$\\ 
% \midrule
% UPP & $.556$& $.348$& $.532 \pm.01$& $.330 \pm.01$& $.533$& $.331$& $.540 \pm.01$& $.342 \pm.01$& $.411$& $.249$& $.408 \pm.01$& $.248 \pm.01$ \\ 

% FT & $.503$& $.325$& $.380 \pm.03$& $.232 \pm.02$& $.517$& $.330$& $.402 \pm.03$& $.250 \pm.02$& $.421$& $.260$& $.349 \pm.02$& $.212 \pm.01$ \\ 
% ER & $.491$& $.314$& $.403 \pm.02$& $.247 \pm.01$& $.521$& $.331$& $.416 \pm.03$& $.257 \pm.02$& $.429$& $.263$& $.352 \pm.02$& $.213 \pm.01$ \\ 
% EWC & $.483$& $.299$& $.445 \pm.03$& $.271 \pm.02$& $.475$& $.294$& $.465 \pm.02$& $.287 \pm.02$& $.429$& $.262$& $.356 \pm.02$& $.214 \pm.01$ \\ 
% TIE & $.548$& $.354$& $.384 \pm.03$& $.224 \pm.02$& $.567$& $.362$& $.417 \pm.03$& $.251 \pm.02$& $.492$& $.309$& $.313 \pm.03$& $.186 \pm.02$ \\ 
% FULL & $.565$& $.358$& $.453 \pm.02$& $.273 \pm.02$& $.555$& $.349$& $.467 \pm.03$& $.286 \pm.02$& $.416$& $.256$& $.363 \pm.02$& $.221 \pm.01$ \\ 
% UPP & $.556$& $.348$& $.530 \pm.01$& $.328 \pm.01$& $.533$& $.331$& $.541 \pm.01$& $.343 \pm.01$& $.411$& $.249$& $.408 \pm.01$& $.248 \pm.01$ \\ 

        \bottomrule
    \end{tabular}
    
\caption{Performance comparison (Hit@10 and MRR) of the RE-NET model incrementally trained using three benchmarks: ICEWS-M, ICEWS-2M, and GDELT. Performance is evaluated at the final training time step over the last test dataset(Current) and across all prior test datasets (Average).}
\label{tab:results}
\vspace{-0.5cm}
\end{table*}
Additionally, we train an upper-bound model, denoted as \textbf{UPP}. During the $t$-th step of training, this model has access to all training data from all preceding time steps, $1, \dots, t$. Detailed information about hyperparameter selection and implementation is provided in Appendix \ref{sec:hyperparams}. The results of this experiment, summarized in Fig. \ref{fig:fig:main}, demonstrate that our full training framework outperforms all other incremental training strategies in alleviating catastrophic forgetting. The L2 regularization used with TIE proves to be overly restrictive, leading to an even greater performance drop than that observed with the finetuning strategy. Table \ref{tab:results} summarizes the performance of the model at the final training time step on the last test dataset (referred to as 'current'), as well as its average performance across all previous test datasets (referred to as 'average'). Despite a slight dip in performance on the current task, our method consistently delivers a higher average performance. This discrepancy underscores the trade-off inherent in our approach, which is deliberately calibrated to strike a balance between maintaining high performance across all tasks and mitigating the forgetting of prior tasks.

\subsection{Ablation Study} 
\label{sec:ablation}
\begin{figure}[t!]
    \centering
    \includegraphics[width=.48\textwidth]{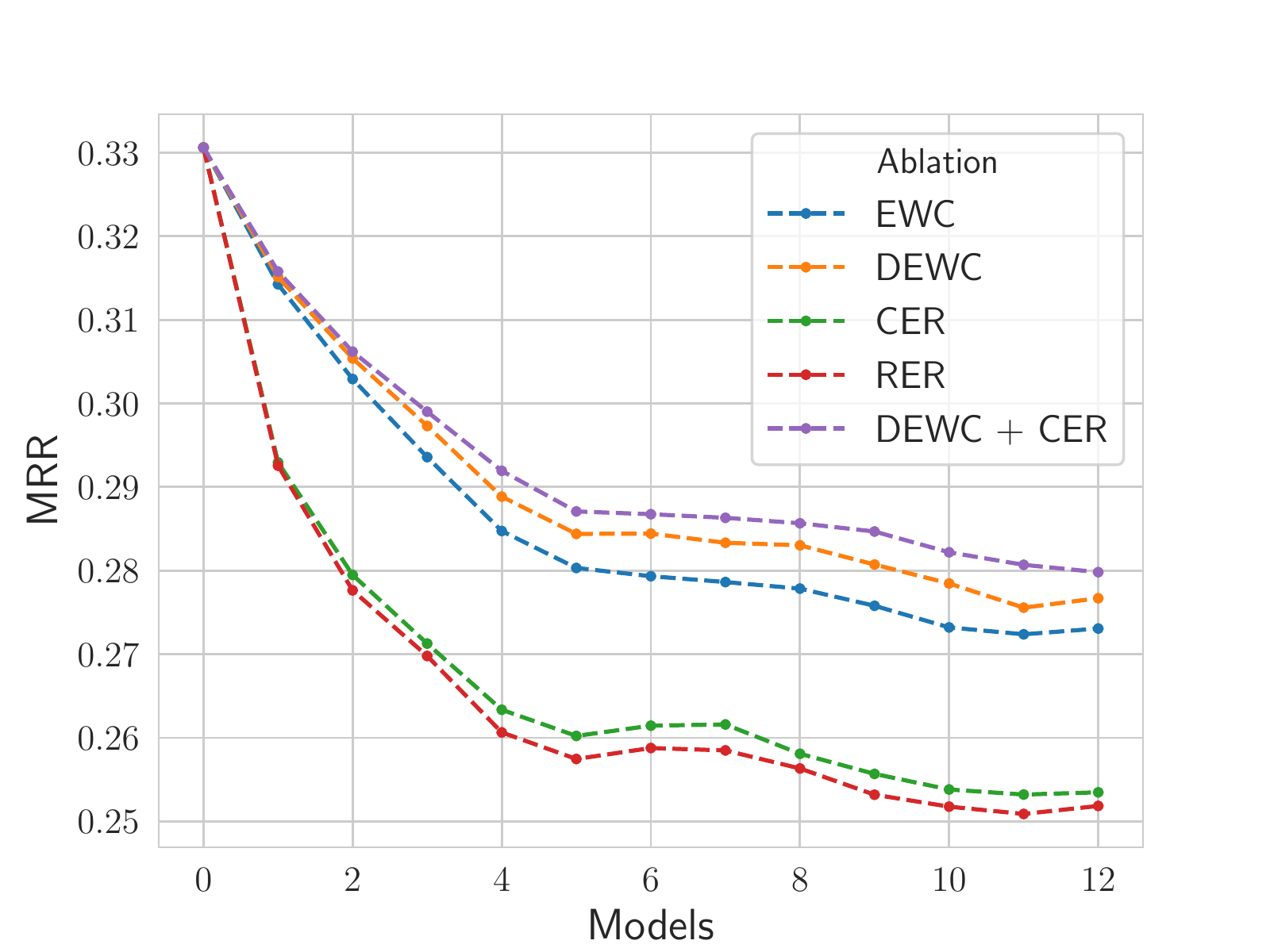}
    \caption{Ablation study on different components of the model using \datam. RER and CER stand for random and clustering-based experience replay. DEWC is the EWC with decayed $\lambda$ values.}
    \label{fig:model_ablation}
   
\end{figure}

In this section, we present an ablation study to evaluate the effectiveness of our proposed approach. Fig. \ref{fig:model_ablation} illustrates the results of various variations of our model, trained on \datam and evaluated using average MRR as the performance metric. The variations include: (1) Random Experience Replay (RER), where points are randomly sampled uniformly; (2) Clustering-based Experience Replay (CER), where points are sampled using the method described in Section \ref{sec:cer}; (3) Regular EWC outlined in Equation \ref{eq:temp_ewc} (EWC); (4) Decayed Elastic Weight Consolidation (DEWC), using the decayed $\lambda$ value outlined in Equation \ref{eq:temp_ewc_lambda}; and (5) DEWC + CER, which represents our full model.

Our results demonstrate that the individual components of our model play a role in enhancing the overall performance, with clustering-based experience replay showing superior performance compared to random experience replay. Additionally, the decayed EWC technique proves to be more effective than the traditional EWC when tasks are assigned equal importance coefficients. For a more in-depth understanding, the detailed results for all datasets used in the ablation study are provided in the Appendix \ref{sec:appendix_ablation}.

\subsection{EWC Variations}
\begin{figure}[t!]
    \centering
    \includegraphics[width=.48\textwidth]{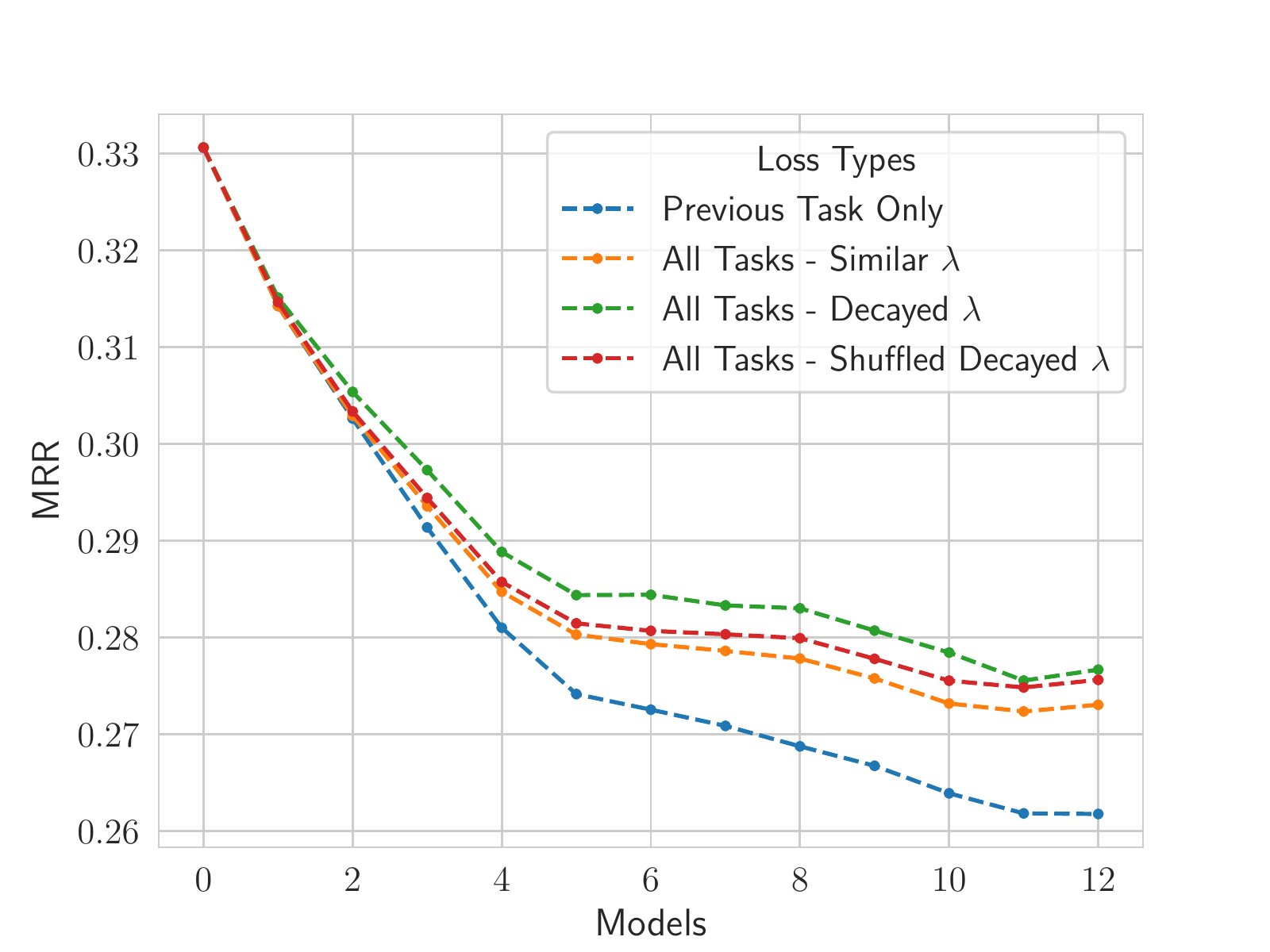}
    \caption{Comparison of EWC loss variations on model performance. Blue line represents using only the previous task in EWC loss, showing a significant reduction compared to considering all tasks. }
    \label{fig:ewc}
\end{figure}

In order to demonstrate the effectiveness of the EWC loss with weight decay (as outlined in Equation \ref{eq:temp_ewc_lambda}), we are comparing it against three other variations of the EWC loss. We will train the \renet method incrementally, using each variation of the EWC loss separately. The results of this comparison can be seen in Fig. \ref{fig:ewc}, which shows the average MRR score for a model trained incrementally with each loss variation, using the \datam dataset. The other variations of the EWC loss that we are comparing against include: (i) only using the parameters of the previous task for regularization, and only computing the Fisher Information Matrix for the previous task; (ii) using all previous task parameters for regularization, but giving all tasks the same importance coefficient value $\lambda$, and computing the Fisher Information Matrix for each task separately (as outlined in Equation \ref{eq:temp_ewc}); and (iii) a variation similar to the second one, but with the decayed $\lambda_i$ values of Equation \ref{eq:temp_ewc_lambda} being assigned to each task randomly. The results in Fig. \ref{fig:ewc} indicate that using only the parameters of the previous task for regularization performs the worst. Using the same $\lambda$ value for all tasks has a smoothing effect on the Fisher Information Matrix, and this is why the decayed, permuted $\lambda$ values perform better. Our proposed loss ultimately outperforms all variations, highlighting the importance of more recent tasks compared to older tasks. As a potential next step, we could investigate learning $\lambda$ values based on task similarities.

\subsection{Memory size and Experience Replay}
\begin{figure}
    \centering
    \includegraphics[width=.48\textwidth]{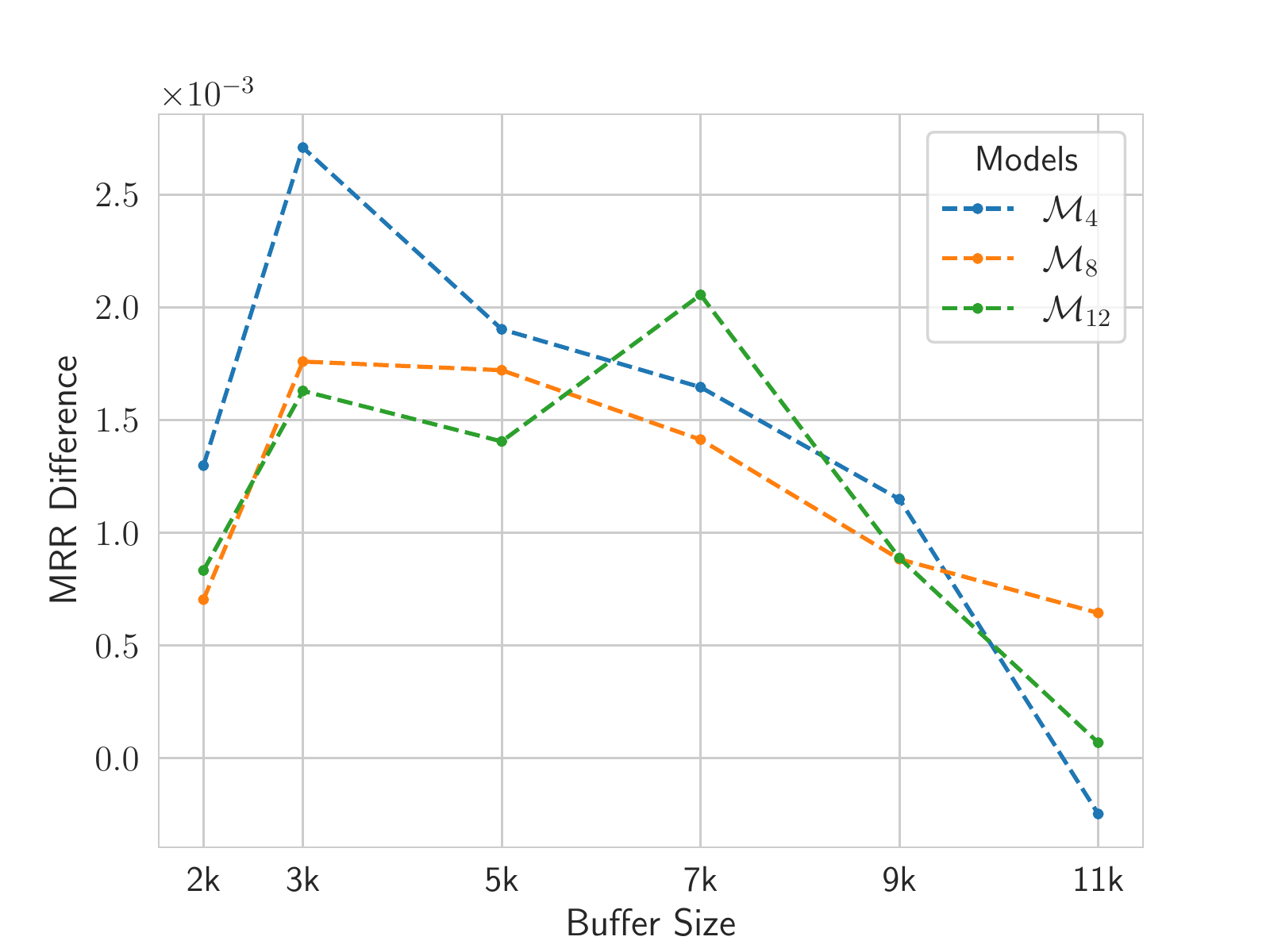}
    \caption{Comparison of average MRR for CER and RER. Results show no significant difference when memory is very small or large, but CER is more effective with sufficient memory.}
    \label{fig:buffer}
\end{figure}

This experiment compares the effectiveness of clustering-based sampling and uniform sampling for experience replay when memory is limited. We use \datam and run \renet with two types of experience replay: (i) random (uniform) sampling (RER) and (ii) clustering-based sampling (CER) using buffer sizes from 2000 to 11000 data points. We evaluated the model performance for $\calM_4$, $\calM_8$, and $\calM_{12}$ which were trained incrementally with experience replay up to time 4, 8, and 12, respectively. We measure the performance of the model by taking the average MRR score over the first $4, 8, 12$ test sets for $\calM_4, \calM_8, \calM_{12}$ respectively. Finally, we compare the performance of RER and CER methods by subtracting the RER model performance from the CER model performance, and the results are shown in Fig. \ref{fig:buffer}. The results, shown in Fig. \ref{fig:buffer}, indicate that when memory is very small or very large, there is no significant difference between RER and CER methods; when memory is too small, there is not enough information for the model to have a significant impact on performance, and when memory is too large, important data points are likely to be selected at random. However, when memory is sufficient, clustering-based sampling becomes more important.

\section{Conclusion}
We propose a framework for incrementally training a TKG completion model that consolidates the previously learned knowledge while capturing new patterns in the data. Our incremental learning framework employs regularization and experience replay techniques to alleviate the forgetting problem. Our regularization method is based on temporal elastic weight consolidation that assigns higher importance to the parameters of the more recent tasks. Our selective experience replay method uses clustering over the representation of the data points and selects the data points that best represent the underlying data structure. Our experimental results demonstrate the effectiveness of our proposed approach in alleviating the catastrophic forgetting for the event-centric temporal knowledge graphs. This work is the first step towards incremental learning for event-centric knowledge graphs. Potential future work might involve exploring, and taking into consideration the effect of time on task similarities which might differ for various applications. 
\section{Limitations}

In this section, we examine the limitations of our approach. Even though our training methodology runs faster and uses less memory than retraining, there remains potential for further scalability optimization. One potential avenue for improvement could involve optimizing the estimation of the Fisher Information Matrix. Furthermore, optimizing the parameters related to the incremental training such as buffer size and regularization coefficient is dependent on the entire time steps rather than the current time steps. Devising a time-efficient way for hyperparameter optimization could be extremely beneficial for this task. Additionally, while our full model has demonstrated some mitigation of the problem of catastrophic forgetting, a significant gap remains between the upper performance bound and the performance of our approach. Further research is necessary to bridge this gap and improve overall performance. Finally, our current focus on continual learning is limited to the emergence of new events and does not currently consider the possibility of new relations or entities. This limitation is in part due to the base model (\renet) not being inductive and is a problem that is inherent to the model itself. Future research in the field of continual learning may aim to address this limitation by considering new relations and entities, even in the context of base models that do not support these features.

\bibliography{ref}
\bibliographystyle{acl_natbib}

\clearpage
\appendix
\begin{figure*}[ht]
    \centering
    \includegraphics[width=\textwidth, trim={3cm 0 3cm 0}, clip]{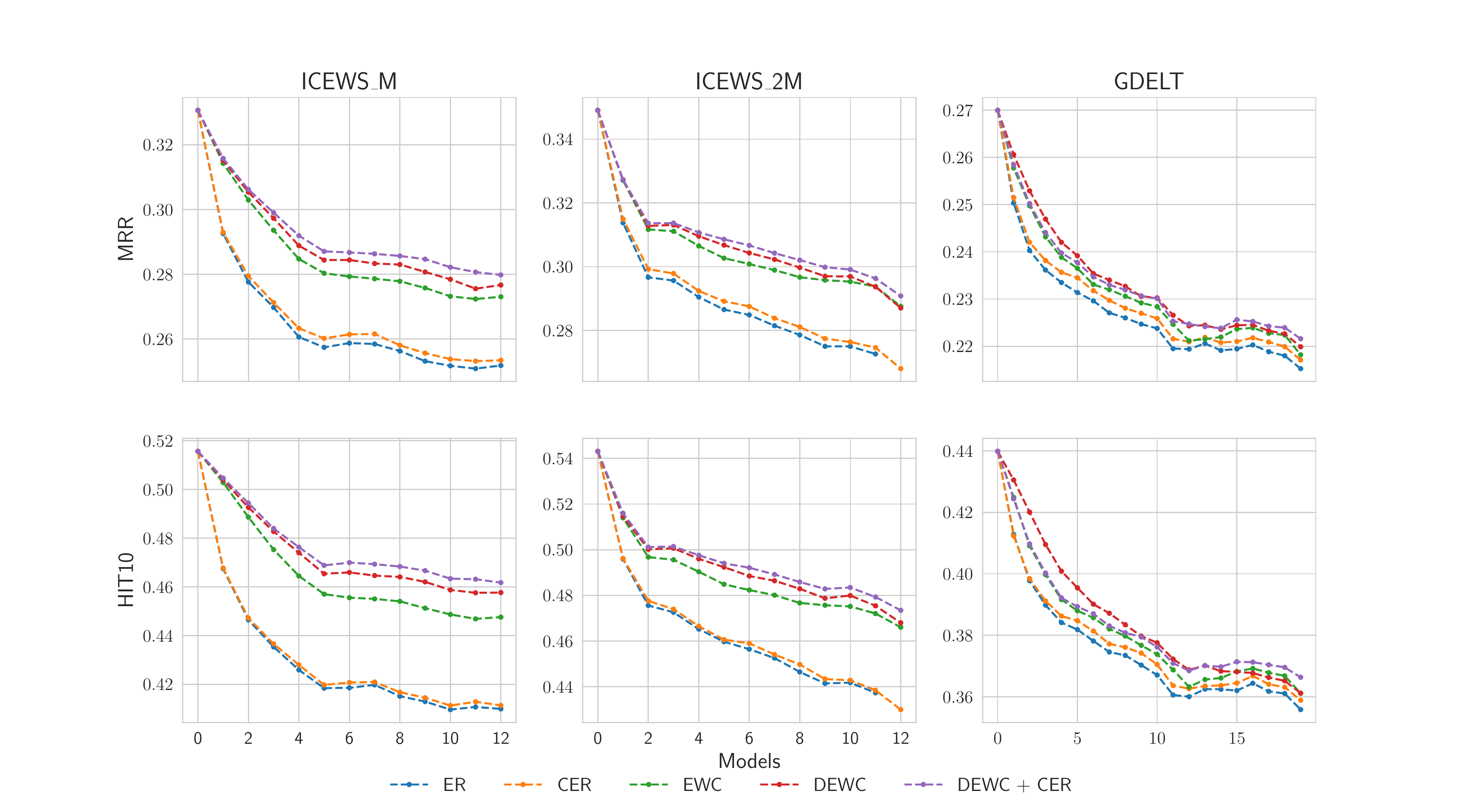}
    \caption{Ablation study on different components of the model using \datam, \datatm, and \gdelt. RER and CER stand for random and clustering-based experience replay. DEWC is the EWC with decayed $\lambda$ values.}
    \label{fig:full_ablation}
    \vspace{-.5cm}
\end{figure*}
\section{Implementation Detail \& Hyperparameters}
\label{sec:hyperparams}
We implemented our models using PyTorch, utilizing the \renet implementation from their GitHub repository\footnote{https://github.com/INK-USC/RE-Net.git} as a base. We modified the training pipeline of \renet and added experience replay and regularization loss. The \renet model utilized a mean pooling layer for the neighborhood encoder, with a dropout of $0.5$ and an embedding dimension of $100$ for relations and entities. For the model variation that employed only EWC loss, we set the learning rate to $10^{-3}$. The regularization coefficient for EWC is set to $10$ and the weight decay to $0.9$ for all the datasets. For variations that included experience replay buffer or fine-tuning, we began training with a learning rate of $10^{-3}$ and decreased it to $10^{-4}$ for subsequent time steps. The buffer size was set to $3000$ for \datam and \gdelt and $5000$ for \datatm, and the batch size was $256$ for \datam and \gdelt and $512$ for \datatm. We selected the best model using the validation set at each time step. We ran each experiment once for each set of hyperparameters as the \renet performance did not vary significantly between runs. The min cluster size for \hdb is set to 5 for all three datasets. We run all the experiments on machines with \texttt{NVIDIA GeForce RTX 2080 Ti} GPUs.

\section{Extended Ablation Study}
\label{sec:appendix_ablation}
In this section, we present the results of the ablation study conducted in Section \ref{sec:ablation} to evaluate the effectiveness of our method. Fig. \ref{fig:full_ablation} illustrates various variations of our model, which were trained incrementally over \datam, \datatm and \gdelt using the hyperparameters reported in the previous section. The model variations include (1) Random Experience Replay (RER), where points are randomly sampled uniformly; (2) Clustering-based Experience Replay (CER), where points are sampled using the method described in Section \ref{sec:cer}; (3) Regular EWC outlined in Equation \ref{eq:temp_ewc} (EWC); (4) Decayed Elastic Weight Consolidation (DEWC), using the decayed $\lambda$ value outlined in Equation \ref{eq:temp_ewc_lambda}; and (5) DEWC + CER, which represents our full model. The results indicate that clustering-based experience replay outperforms random experience replay, and that the DEWC approach is more effective for the ICEWS datasets compared to GDELT. This may be due to the fact that the data distribution for ICEWS datasets changes more significantly over the course of a year compared to GDELT, which only includes 21 days of data. It is also visible from the plots that the GDELT dataset exhibits less forgetting compared to both ICEWS datasets. Finally, the full model (DEWC + CER) always outperforms the other model variations, demonstrating the effectiveness of our methodology.

\end{document}